\documentclass{article}

\usepackage{PRIMEarxiv}

\usepackage[utf8]{inputenc} % allow utf-8 input
\usepackage[hidelinks]{hyperref}       % hyperlinks
\usepackage{url}            % simple URL typesetting
\usepackage{booktabs}       % professional-quality tables
\usepackage{amsfonts}       % blackboard math symbols
\usepackage{nicefrac}       % compact symbols for 1/2, etc.
\usepackage{microtype}      % microtypography
\usepackage{lipsum}
\usepackage{fancyhdr}       % header
\usepackage{graphicx}       % graphics
\graphicspath{{media/}}     % organize your images and other figures under media/ folder

\usepackage{amssymb}
\usepackage{amsmath}
\usepackage{multirow}
\usepackage{subfig}

\title{Data-Driven Short-Term Daily Operational \\ Sea Ice Regional Forecasting
}

\author{
  Timofey Grigoryev\thanks{Correspondence: \texttt{timofey.a.grigoryev@gmail.com}}, \;
  Ilya Trofimov, \;
  Nikita Balabin, \;
  Evgeny Burnaev, \;
  Vladimir Vanovskiy \\
  Applied AI Center \\
  Skolkovo Institute of Science and Technology \\
  Moscow, Russia\\
  \texttt{\{t.grigorev, ilya.trofimov, nikita.balabin, e.burnaev, v.vanovskiy\}@skoltech.ru} \\
  \And
  Polina Verezemskaya, \;
  Mikhail Krinitskiy, \;
  Alexander Gavrikov, \;
  Sergey Gulev \\
  Shirshov Institute of Oceanology \\
  Moscow, Russia\\
  \texttt{\{verezem, krinitsky, gavr, gul\}@sail.msk.ru} \\
  \And
  Nikita Anikin \\
  Moscow Institute of Physics and Technology \\
  Dolgoprudny, Russia\\
  \texttt{anikin.nn@phystech.edu} \\
  \And
  Aleksei Shpilman, \;
  Andrei Eremchenko \\
  Gazprom Neft \\
  St.~Petersburg, Russia\\
  \texttt{\{Shpilman.AA, Eremchenko.AYu\}@gazprom-neft.ru}
}

\begin{document}
\maketitle

\begin{abstract}
Global warming made the Arctic available for marine operations and created demand for reliable operational sea ice forecasts to make them safe. While ocean-ice numerical models are highly computationally intensive, relatively lightweight ML-based methods may be more efficient in this task. Many works have exploited different deep learning models alongside classical approaches for predicting sea ice concentration in the Arctic. However, only a few focus on daily operational forecasts and consider the real-time availability of data they need for operation. In this work, we aim to close this gap and investigate the performance of the U-Net model trained in two regimes for predicting sea ice for up to the next 10 days. We show that this deep learning model can outperform simple baselines by a significant margin and improve its quality by using additional weather data and training on multiple regions, ensuring its generalization abilities. As a practical outcome, we build a fast and flexible tool that produces operational sea ice forecasts in the Barents Sea, the Labrador Sea, and the Laptev Sea regions.
\end{abstract}

% keywords can be removed
% \keywords{First keyword \and Second keyword \and More}

\section{Introduction}

The rapid Arctic warming \cite{screen2010central} is characterized by a twice as substantial temperature increase compared to the global mean \cite{arndt2015state, blunden2016state, hausfather2017assessing}. According to the ERA5 reanalysis, the annual Arctic warming trend from 1979 to 2020 is 0.72 ℃/decade \cite{you2021warming}, which is 2-3 times stronger than the global mean.
Arctic rapid warming is closely associated with an unprecedented decline of sea ice extent by more than 30\% over the last four decades \cite{kwok2018arctic, perovich2021meltwater}, and a decrease of sea ice thickness \cite{renner2014evidence}. These changes allow for faster and cheaper sea routes, such as the Northeast Passage \cite{williams2021presentation}. Sea ice jams are one of the most critical problems in marine navigation security. Accurate operative forecasts of sea ice properties and dynamics can mitigate that problem, allowing ships to adjust their routes to avoid regions of ice accumulation. At the same time, new routes through the Arctic will cause an increase in the ocean and atmospheric pollution risks, primarily due to fishing, oil/gas extraction, and transportation. For the delivery of natural gas and oil to long-distance destinations, transport by deep-sea vessels is more economical compared to offshore pipelines \cite{aseel2021model}.
% Oil and gas tankers are primarily powered by marine diesel and heavy fuel oil, highly damaging to the sophisticated marine Arctic nature.
To decrease ocean pollution and the carbon footprint \cite{greene2020well, ankathi2022greenhouse} caused by transportation, gas/oil companies must optimize the routes \cite{kaleschke2016smos} to make them faster and to reduce associated ecological risks (for example, reduce the atomic icebreaker usage).

Coupled ocean-ice numerical modeling is the evident source of a reliable forecast of sea ice conditions. Newest sea ice models, such as NextSIM \cite{rampal2016nextsim, williams2021presentation} demonstrate fascinating results on sea ice concentration, thickness and drift vectors representation comparing to the observational data (OSI SAF SSMI-S \cite{tonboe2016product}, AMSR2 \cite{lavelle2016product}, GloblICE dataset, \url{http://www.globice.info}). NextSIM is a fully-Lagrangian finite-element model, making it tough to couple with Euler method-based ocean models. Eulerian sea ice models have been evolving for the last two decades and can reproduce some aspects of sea ice and its recent changes. However, detailed comparisons between satellite remote sensing data with Eulerian-model results reveal big differences in certain aspects of the sea ice cover, e.g., for fracture zones and small-scale dynamic processes \cite{kwok2008icesat, girard2009evaluation}. It remains unclear whether the current model physics (elastic-viscous-plastic rheology) is suitable for reproducing these observed sea ice deformation features \cite{bouillon2015presentation, girard2011new, sulsky2007using} and provides a reliable forecast. Furthermore, coupled ocean-ice numerical modeling requires significant computational resources.

Statistical or data-driven machine learning approaches, on the other hand, are more flexible and lightweight. They do not need a complex physical model of processes in the ocean and atmosphere to work. Once trained, such a model only needs appropriate recent observations and comparatively little computational resources to make a forecast. However, the training part in this case is quite difficult for several reasons. First, most of the input data used for training (including sea ice concentration) is presented as 3d or even 4d spatiotemporal maps with a huge amount of highly correlated input channels. It has been found, that usage of modern convolutional \cite{lecun1989conv, lecun1990conv, lecun1998conv, ronneberger2015unet, he2016resnet}, recurrent \cite{hochreiter1997long, cho2014learning} or attention-based \cite{google2017attention, google2015spatial} architectures can overcome difficulties associated with exploding number of trainable parameters and overfitting.
Second, the model’s output is expected to be a consistent SIC forecast retaining the same spatiotemporal nature, which is hard to guarantee when training on a limited amount of data. In order to overcome these difficulties, one can train a model not to predict the data itself but to compensate for the errors of simple baselines, such as climatology mean, persistence, or cell-wise linear trend.
% However, a naive approach to treating this problem as a regression with standard $L_p$ loss can lead to averaged blurry predictions. One should use more sophisticated techniques to overcome these difficulties, such as predicting probability distribution over possible answers, exploiting adversarial training, or embedding physical principles into the models.
Finally, operative climate and sea ice characteristics data have their peculiarities. It usually consists of several patches obtained at different times each day, thus should be combined and averaged daily. SIC can only be measured in the sea, leaving the land cells blank. Measurements can be based on different sources inheriting different biases, making the signal-to-noise ratio lower than expected. Furthermore, the actual changes in the sea ice condition occur in limited periods in fall and spring, making more than half of the data barely usable. Considering everything above, one must be very thoughtful when designing training and testing pipelines and choose proper metrics to assess obtained solutions adequately.

Many works are dedicated to sea ice forecasting in the Arctic region. However, research in this field mainly focuses on climate studies rather than operative sea ice forecasts for practical use.
Fully-connected MLP is often used either as the primary method for predicting monthly-averaged sea ice concentration \cite{kim2019satellite} or as one of the benchmarks \cite{kim2017prediction, wang2017sicestimation}. NSIDC Nimbus-7 SMMR and DMSP SSMI/SSMIS data are used there as SIC maps. Other approaches exploit CNN, applied on patches cropped out of ice maps \cite{wang2017sicestimation}, or RF with an additional set of weather input features from ERA-Interim \cite{kim2020prediction}. Deep learning methods are compared with simpler baselines in these works and reported to perform significantly better in standard metrics, such as RMSE.
Works \cite{liu2021extended, andersson2021icenet} are of particular interest, as they consider more advanced deep learning models that seem more suitable for sea ice forecasting.
In \cite{liu2021extended} authors consider ConvLSTM \cite{shi2015convlstm} model, which can fully make use of spatial-temporal structure of the climatological data. However, they use weather maps (predictors) from ERA-Interim and ORAS4 NEMO reanalysis data for training, thus limiting model applicability for operational sea ice forecasts. Authors evaluate the performance of ConvLSTM on a weekly-averaged and monthly-averaged scale and obtain results comparable in terms of RMSE to those of the ECMWF numerical climate model only for short lead times.
Authors of \cite{andersson2021icenet} deal with U-Net \cite{ronneberger2015unet} model and train it to predict probabilities for the next 6 months for monthly-averaged SIC values in each cell to belong to each of three classes: open water, marginal ice and packed ice. They thoroughly investigate the model properties and compare it with SEAS5, a numerical ocean-ice model with state-of-the-art sea ice prediction skills. However, the paper does not consider possibilities for operating at the daily temporal resolution.

In our work, we focus on the operative daily sea ice forecasting and imply corresponding restrictions on the weather and sea ice data we use. To our knowledge, only a few papers consider this type of setting. However, all these works either use non-operative reanalysis data or perform experiments with one or two currently outdated machine learning methods.
For example, \cite{fritzner2020assessment} demonstrates the potential of machine learning in sea ice forecasting by comparing a numerical ocean-ice model with simple CNN and cell-wise k-NN method. Unlike previous works, it focuses on short-term predictions with a length of 1-4 weeks.
\cite{choi2019ann} assesses the ability of different cell-wise GRU networks equipped with feed-forward encoder and decoder to forecast SIC for up to the next 15 days. To overcome limitations of locality in this setting, the authors incorporate global statistics in the network inputs and report significant improvement in the prediction accuracy.
Authors of \cite{liu2021dailyprediction} demonstrate the superiority of ConvLSTM over CNN when forecasting SIC data in a patch-wise manner with patches of size 41 by 47 pixels. They use only NSIDC Nimbus 7 and DMSP SMMR SIC data, which is available operatively but has a low resolution (25 $\times$ 25 km) to be of actual use in navigation, and forecast daily-averaged SIC for the next 10 days.
In \cite{liu2021shortterm} authors investigate variations of relatively modern PredRNN++ \cite{wang2018predrnn} architecture for SIC forecasting for the next 9 days and compare it to the ConvLSTM network, demonstrating the superiority of the former. However, their model depends on ECMWF ERA5 reanalysis data, which is not available in real-time, and thus limits its practical value.

In this work, we provide a thorough research on the prospects of machine learning in sea ice forecasting in a few regions in the Arctic: the Barents and Kara Seas (Barents), the Labrador Sea (Labrador), and the Laptev Sea (Laptev). These three regions demonstrate varying SIC inter-annual dynamics and allow the investigation of the model's performance in different conditions. We deal with SIC and weather data that is available in real-time and can be used in practice to obtain operational SIC forecasts for marine navigation. A single simple yet effective classical U-Net neural architecture is chosen as such a model. It is lightweight, thus not prone to overfitting, and suited well for image-to-image tasks, such as sea ice forecasting. We treat JAXA AMSR-2 Level-3 imagery as ground truth of sea ice concentration maps and train, validate and test our models on this data. As a result, we not only obtain a trained U-Net model for the operational sea ice forecasts but also provide datasets we used for benchmarks and future comparisons for the research community.
All similar works test their models with different satellite data in different regions during different periods over varying baselines and usually report improvement in MAE in the range 25\% -- 50\% over considered baselines. Though the comparison with them hardly makes sense, we obtain similar daily improvement over persistence around 25\% in all three regions.

The main contributions of our work are the following:
\begin{enumerate}
    \item We collect JAXA AMSR-2 Level-3 SIC data and GFS analysis and forecasts data, process it and construct three regional datasets, which can be used as benchmark tasks for future research.
    \item We conduct numerous experiments on forecasting SIC maps with the U-Net model in two regimes and provide our findings on the prospect of this approach, including comparison with standard baselines, standard metric values, and model generalization ability.
    \item We build a fast and reliable tool --- trained on all three regions U-Net network that can provide operational sea ice forecasts in these Arctic regions.
    \item We compare U-Net performance in forecasting in recurrent (R) and straightforward (S) regimes and highlight the strength and weaknesses of both.
\end{enumerate}

\section{Data}

\subsection{Sea Ice Data (JAXA AMSR-2 Level-3)}

Plenty of sea ice concentration products are available, covering a period from the very beginning of the satellite era to nowadays \cite{kern2020satellite, kern2019satellite, cavalieri200330}. Most of them are available daily on the regular grid with spatial resolution varying from 12.5 to 50 km \cite{kern2020satellite}. We were looking for a higher resolution satellite product to provide a forecast comparable to high-resolution ocean sea ice modeling. We present our analysis of the sea ice conditions based on JAXA (\url{http://ftp.eorc.jaxa.jp}) as a High-Resolution Sea Ice Concentration Level-3 from Advanced Microwave Scanning Radiometer-2 (AMSR2 hereafter) from the GCOM-W satellite. Daily sea ice concentration is available since July 2, 2012, with a spatial resolution of 5 km on the regular grid, which is the highest resolution compared to other SIC datasets, that we could find in openly available products (CRYOSAT, AMSR2 L4, SSMI, and other). SIC data is given in percentages (\%) from 0 to 100. We used data from July 2, 2012, to January 20, 2022, i.e., 6970 days. AMSR2 L3 research product of SIC is distributed in two daily entities corresponding to the composites combined from the data acquired during ascending and descending satellite passes. In our study, we use the mean between these two daily snapshots as the ones statistically closer to ground truth compared to individual composites.

\begin{figure}[!htbp]
    \includegraphics[width=\textwidth]{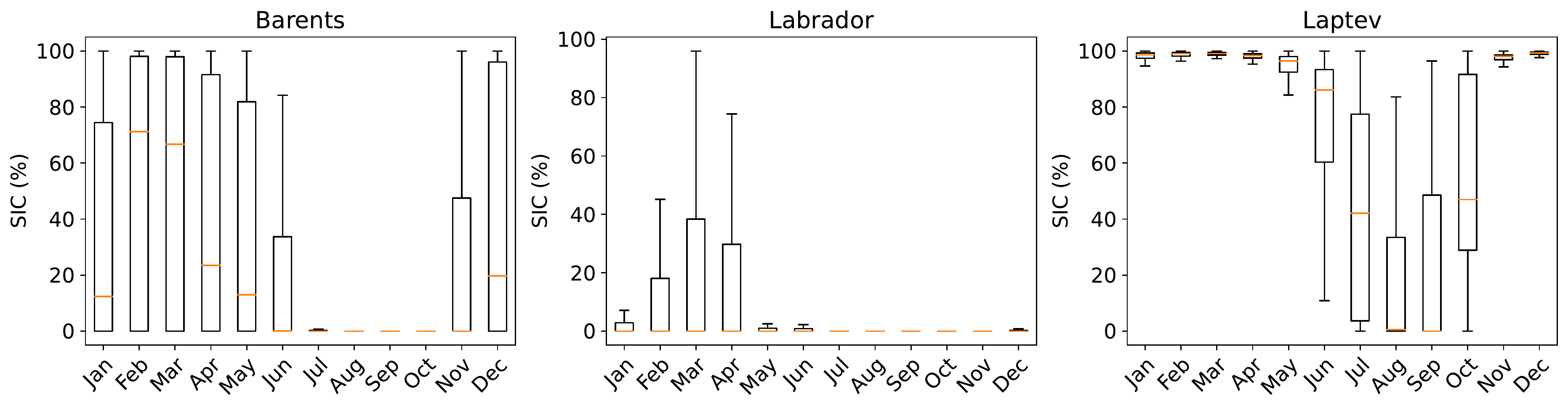}
    \centering
    \caption{Box and whisker plots of SIC data distribution in JAXA for different months of 2021, aggregated for all the cells in each region. The box extends from the 25th percentile to the 75th percentile; whiskers extend the box by 1.5x of its length. The orange line is the median (50th percentile); outliers are omitted in order not to clutter the plot.}
    \label{jaxa_sic-data-month}
\end{figure}

\begin{figure}[!htbp]
    \includegraphics[width=\textwidth]{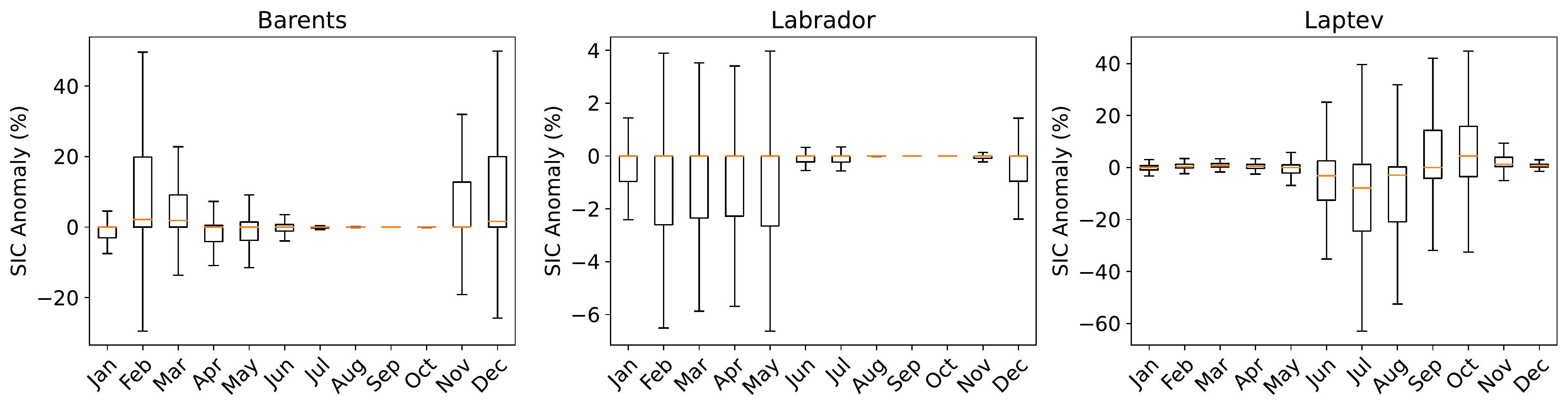}
    \centering
    \caption{Box and whisker plots of SIC climatological anomaly distribution in JAXA for different months of 2021, aggregated for all the cells in each region. Climatological anomaly is the difference between the data and the climatology of the respective channel (see subsection \ref{section:data_preprocessing}). The box extends from the 25th percentile to the 75th percentile; whiskers extend the box by 1.5x of its length. The orange line is the median (50th percentile); outliers are omitted in order not to clutter the plot.}
    \label{jaxa_sic-anomaly-month}
\end{figure}

Monthly statistical distributions of SIC are presented in figure \ref{jaxa_sic-data-month} and its climatological anomalies in figure \ref{jaxa_sic-anomaly-month} (see subsection \ref{section:data_preprocessing} for details of its computation). Changes in sea ice are present only in about half of the months during the year. In the remaining time, the regions are fully melted (Barents, Labrador) or fully frozen (Laptev).

\subsection{Weather Data (GFS)}

While sea ice concentration describes its condition and dynamics, there is an opportunity for potential improvement of a statistical model using additional variables correlated with sea ice dynamics. For example, surface winds influence sea ice drift, especially in shallow seas. Surface air temperature may also impact sea ice dynamics through ice melting or growth. Our study explored the potential for improving data-driven SIC forecast by extending input features with atmospheric properties such as 2-meter temperature, surface pressure, and u, v components of wind and its absolute speed.

We used NCEP operational Global Forecast System (GFS) for atmospheric and ice condition data. The GFS core is based on coupled atmospheric-ocean-ice models and provides an analysis and forecast globally at 0.25$^{\circ}$ horizontal resolution and 127 vertical levels (for atmosphere) \cite{GFS2015}. Model forecast runs up to 16 days in advance at a 3 hourly time steps interval at 00, 06, 12, and 18 UTC daily. The output is openly available in WMO GRIB2 format. It is available with no delay and a minimal number of temporal gaps, making it the best choice between the weather data sources for a reliable operative forecasting system.

\subsection{Regions}
\label{section:regions}

\begin{figure}[!htbp]
    \includegraphics[width=\textwidth]{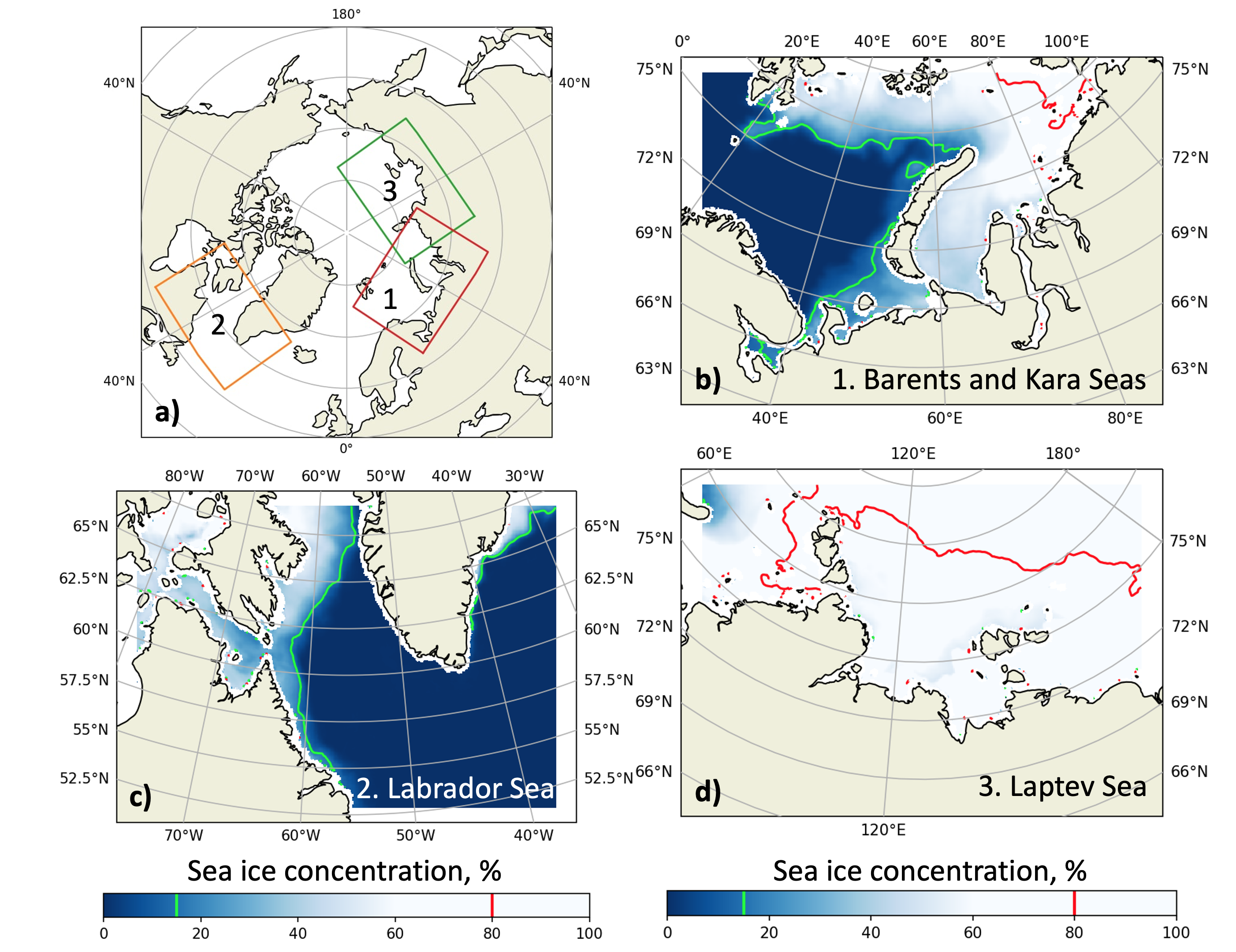}
    \centering
    \caption{Regions definitions. (a) Arctic with colored boxes indicating the three regions: 1, the Barents and Kara Seas (b); 2, the Labrador Sea (c); and 3, the Laptev Sea (d). Mean sea ice concentration (\%) and 15\% (green) and 80\% (red) concentration isolines are drawn for 2021. The area between the green and red isolines is the marginal ice zone.}
    \label{sicmap_all}
\end{figure}

We conduct experiments on three regions with varying SIC inter-annual dynamics (figure \ref{sicmap_all}). It allows us to enlarge the dataset and to adjust the model for different sea ice conditions. The Labrador Sea presents the Atlantic type of ice regime, characterized by the shortest period of pack ice in the basin (1-3 months) with mean SIC below 50\% during the coldest month in a year. High interannual variability of the SIC in the Labrador Sea is caused by the sea ice fragmentation observed in the marginal ice zone (15-80\%, MIZ hereafter) due to the ocean-wave-ice-atmosphere interaction. The Laptev Sea shows the typical inner-Arctic icing: SIC is above 90\% spanning 7-9 months in the annual sea ice duration with strong sea ice freeze-up and slight sea ice freeze-up. The highest inter-annual variability is observed in summer, resulting in large open-water areas (SIC $\ll$ 80\%). The Barents Sea and the Kara Sea regions are a mixture of these two types. The Barents Sea is an ice-free region due to the influence of intense warming from the North Atlantic Current. On the other hand, the Kara Sea is separated by the island of Novaya Zemlya and is similar to the Laptev Sea.

To prepare the dataset, we projected the regions of interest onto the new grids, which are the same except for the center points. The projection is described in table \ref{projection_definition}. Center points are described in table \ref{projection_centers}.

\begin{table}[!htbp]
    \centering
    \begin{tabular}{ |c|c|c| } 
    \hline
    \textbf{Parameter Name} & \textbf{Value} & \textbf{Unit} \\
    \hline
    Projection & Lambert Azimuthal Equal Area & - \\
    \hline
    Grid step & 5 & km \\
    \hline
    \multirow{2}*{Grid Height} 
    & 360 & knot \\
    & 1800 & km \\
    \hline
    \multirow{2}*{Grid Width}
    & 500 & knot \\
    & 2500 & km \\
    \hline
\end{tabular}

    \caption{Grid parameters are the same for all the regions.}
    \label{projection_definition}
\end{table}

\begin{table}[!htbp]
    \centering
    \begin{tabular}{ |c|c|c| } 
    \hline
    \textbf{Region} & \textbf{Central Point (Lat, Lon)} & \textbf{Presentation Percentage} \\
    \hline
    Barents & 73\textdegree, 57.3\textdegree & 56.82\%\\
    \hline
    Labrador & 61\textdegree, -56\textdegree & 47.84\%\\
    \hline
    Laptev & 76\textdegree, 125\textdegree & 51.53\%\\
    \hline
\end{tabular}

    \caption{Central points of projections. The presentation percentage is a fraction of satellite data present in each region relative to the region's size (in cells). The lack of data is due to recognition errors and the presence of land (where SIC does not make sense).}
    \label{projection_centers}
\end{table}

\section{Methods}

\subsection{Data Split}

For all the regions, we used data up to the 2019 year for training, 2020 year for validation, and 2021 year for testing. So for models trained just on JAXA, it means around 7.5 years in the training set (since mid-2012); for models trained on both JAXA and GFS, it means around 5 years in the training set (since 2015).

\subsection{Data Preprocessing}
\label{section:data_preprocessing}

It is well known that artificial neural networks train more stably when fed with normalized data \cite{ioffe2015batchnorm}. We compute and use the climatological anomalies instead of raw data for input channels with a strong seasonal cycle, such as GFS temperature and pressure. First, for every channel for every day in a year, we compute climatology. It is an averaged map for that day over all the years in the training set. The averaging is performed within a window of 3 days size for further noise reduction. Then we obtain climatological anomaly for each date by subtracting the climatology of the corresponding day from the raw data. Next, all the channels fed to the model are standardized by linear rescaling, the same for every pixel (but different for different channels). The mean and variance of this transformation are computed over the training set and kept the same for validation and testing sets. Finally, the network outputs (JAXA SIC for all the cases) are rescaled by the inverse transform corresponding to the forecasted channel to be back in the desired range (0 -- 100\%).

\subsection{Baselines}

We consider three types of baselines: persistence, climatology, and trends. Persistence is a constant forecast --- for any day in the future, the value of a parameter in each point is equal to that of today. Climatology baseline forecasts the historical average of a channel for that date over available observations in previous years (see subsection \ref{section:data_preprocessing} for details on its computation). Trends are cell-wise polynomial trends (mean, linear, quadratic, and so on) for values of a parameter for the last $D_\text{in}$ days. We consider trends up to cubic. Only persistence and a 3-day linear trend showed competitive results with the U-Net model. Thus we will report only their metrics for comparison.

\subsection{Models}

We use U-Net network \cite{ronneberger2015unet} in all our experiments. U-Net is originally designed for image segmentation tasks. However, by not applying softmax to the last layer outputs but pixel-wise clipping them to be in the desired range, we adjust it to predict sea ice concentration in the range $[0, 1]$ instead of the logits of the classes probabilities. We exploit the same classical architecture from \url{https://github.com/milesial/Pytorch-UNet} for all the experiments. We only adjust the amount of input and output channels to fit chosen subsets of variables (sea ice and weather maps).

\subsection{Metrics and Losses}

There are three classical metrics in sea ice forecasting, which differ by cell-wise statistic they aggregate: mean absolute error
\begin{equation}
    \label{mae}
    \text{MAE} = \frac{1}{S} \sum_{i,j \in \mathcal{A}} \left| c_{ij}^\text{pred} - c_{ij}^\text{gt} \right| dS_{ij},
\end{equation}
root mean square error
\begin{equation}
    \label{rmse}
    \text{RMSE}^2 = \frac{1}{S} \sum_{i,j \in \mathcal{A}} \left( c_{ij}^\text{pred} - c_{ij}^\text{gt} \right)^2 dS_{ij},
\end{equation}
and integrated ice edge error \cite{goessling2016iiee}
\begin{equation}
    \label{iiee}
    \text{IIEE} = \frac{1}{S} \sum_{i,j \in \mathcal{A}} \left[ \theta(c_{ij}^\text{pred}) \neq \theta(c_{ij}^\text{gt}) \right] dS_{ij}.
\end{equation}
Here indices $i, j$ run over all cells in active subdomain $\mathcal{A}$ of sea ice change. Our study treats all cells with present SIC data as active subdomain cells. Next, $dS_{ij}$ is the area of the respective cell, their sum
\begin{equation}
    S = \sum_{i,j \in \mathcal{A}} dS_{ij}
\end{equation}
is a full area of active subdomain and $\theta (c) = [c > c_0]$ is a threshold function that binarizes SIC value $c$ with threshold $c_0$ and maps it to one of two classes: full ice (1) or open water (0).

Our primary metric is MAE, or pixel-wise $\ell_1$ loss, which is perfectly differentiable. That is why we chose to minimize it during training explicitly. For the RMSE metric, minimizing $\ell_2$ loss can be better. In our experiments, these two losses performed on par. We also considered segmentation setting with two standard classes: open water with SIC $\leq$ 15\% and packed ice with SIC $>$ 15\%. For this setting, we computed the IIEE metric.

\subsection{Augmentations}
\label{section:augmentations}

It is well known that augmentations make training more stable, prevent overfitting and improve the generalization ability of a model \cite{shorten2019augs}. We perform only geometrical transformations: random horizontal flips with a probability of 0.5, rotations on a random angle, uniformly sampled from range $[-30^\circ, 30^\circ]$ (with NaN padding), and translations, uniformly sampled for both vertical and horizontal shifts in the range $[-10\%, +10\%]$ of each dimension relative to map size (with NaN padding). We apply the same transformation to each channel on each input day and compare model output with samely transformed target SIC maps.

\subsection{Regimes}
\label{section:regimes}

\begin{figure}[!htbp]
\centering
    \subfloat[S-regime scheme.]{
        \includegraphics[width=0.4\textwidth]{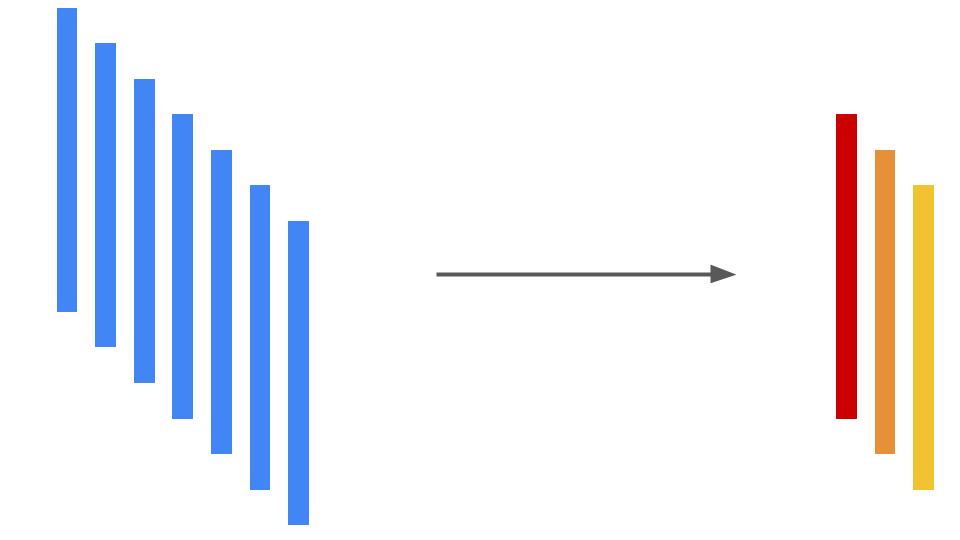}
    }
    \hspace{1cm}
    \subfloat[R-regime scheme.]{
        \includegraphics[width=0.2\textwidth]{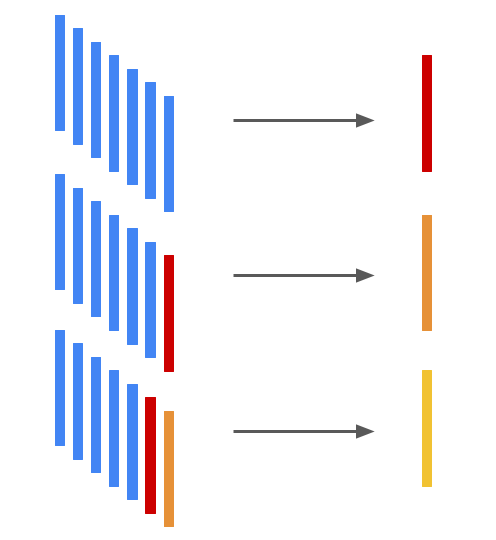}
    }
\caption{Schematic representation of the basic principles of the two U-Net regimes. Blue rods represent blocks of historical data (stacked SIC or weather maps), one for each day in the past. The leftmost is 6 days old, and the rightmost is of today (for 7-day historical input). Red, orange and yellow rods represent model forecasts of SIC maps for the first, the second, and the third days in the future, respectively (for a 3-day forecast). All the forecasts are made simultaneously in S-regime, while in R-regime, they are made one by one, and the model's inputs are updated on each step.}
\label{unet_regimes}
\end{figure}

Classical U-Net from the box is ideally suited to capture spatial correlations in data, but not temporal. In order to overcome this limitation and to avoid the complication of the architecture of the network, we consider two possible regimes to process sequential data with U-Net: a straightforward (S) and a recurrent (R) (see figure \ref{unet_regimes}).

When constructing the network input, we concatenate channels (regional maps of different scalar parameters) from the past (SIC history for the last $D_\text{in}$ days), channels from the future (GFS forecasts for $D_\text{out}$ days), and channels with auxiliary information (latitude, longitude, harmonics of the first date of the forecast and land map). In the straightforward regime, there are $D_\text{out}$ output channels --- one for SIC of each forecast day. In the recurrent mode, we predict only SIC for the next day and use it iteratively as input along with other channels in a recurrent fashion for $D_\text{out}$ times to construct a forecast for $D_\text{out}$ days. However, training that way seems to be subject to the same limitations as any RNN \cite{hoch1998gradprob} and does not go well. Therefore, first we pretrain U-Net to make a forecast for $D_\text{out} = 1$ day and then finetune for $D_\text{out} > 1$ in the described fashion. This approach is inspired by ideas of curriculum learning \cite{bengio2009curiculum} and improves the results significantly compared with no finetuning or no pretraining and allows one to make stable forecasts further in the future.

\subsection{Implementation}

We implemented a training and testing pipeline in Python using the popular machine learning framework PyTorch. We used Adam optimizer with a learning rate $10^{-4}$, learning rate decay rate $10^{-2}$, and batch size of 16 in S-regime and 8 in R-regime. We conducted each experiment on a single NVIDIA A100 40GB GPU, requiring from 1 to 12 hours for training depending on the number of input channels, number of involved regions, and training regime.

\section{Results}

As was mentioned in subsection \ref{section:regimes}, we consider two regimes for time series forecasting using the U-Net backbone: straightforward (S), when the number of output channels is set equal to the number of output days, and recurrent (R), when the number of output channels is set to 1 and the model is iteratively applied for the number of output days times in recurrent fashion. In this section, we will provide the results of conducted experiments for both regimes in different settings and compare them. We conduct experiments on all three regions (Barents, Labrador, Laptev) and present results alongside. The scales of the same diagrams and plots for different regions may vary depending on the specifics of geographical features and sea ice dynamics in each region.

\subsection{Inputs Configuration}

There is a vast number of combinations of possible inputs for the model. JAXA data has just 1 channel --- SIC, but we can not only vary the number of previous days we stack for the model but also the channel's preprocessing. One can compute climatology and climatological anomaly (see subsection \ref{section:data_preprocessing}) and pass it as well, just for the last day or all input days. The same goes for GFS data, from which we could use 5 channels (temperature, pressure, u and v components of wind and its module) and 4 forecasts with a different lead time for each day. On top of that, one can suggest several general channels that can be useful. Firstly, one should consider the harmonics of the current day phase in the year. That is $\cos \varphi$ and $\sin \varphi$ for $\varphi = 2\pi \frac{D}{D_Y}$, where $D$ is a number of a current day from the start of the current year and $D_Y \approx 365.2425$ is the average number of days in a year. It is natural to assume the dynamics of sea ice are different for different seasons, and that input will allow a model to capture these dependencies. Secondly, the binary segmentation map of sea and land may be useful. The region of interest is entirely inside the sea, so it can be useful for the model to treat weather data from the land differently if it falls in the perception core of the convolutions. Thirdly, one can use a map of areas of the grid cells. In our case, all the grid cells are almost the same area due to the choice of the projection type (see subsection \ref{section:regions}). Finally, the grid cells' longitudes and latitudes might be useful too. They can be used for a model trained to perform on one region to better fit parts of it with different climatological properties. On the other hand, they can harm the model's generalization ability and drop its performance in other regions if the values are out of the neural network domain.

It is worth mentioning that stacking all the available channels into the model's input is not the best solution since many of them are highly correlated, and some of them may have no relevant signal for the model. We considered the nature of each channel and conducted a bunch of experiments. It allowed us to choose a single configuration of inputs for both regimes that includes all the available and useful channels. The configuration is described in table \ref{unet_configuration}. For the experiments with no GFS data, we omitted all the channels with source GFS.

\begin{table}[!htbp]
    \centering
    \begin{tabular}{ |c|c|c|c| } 
    \hline
    \textbf{Source} & \textbf{Channel} & \textbf{Preprocessing} & \textbf{Time Interval}  \\
    \hline
    \multirow{1}*{JAXA}
    & SIC & Data & Past \\
    \hline
    \multirow{12}*{GFS}
    & Temperature & Climatology & Today \\ 
    & Temperature & Clim. Anomaly & Today \\ 
    & Temperature & Clim. Anomaly & Future \\ 
    \cline{2-4}
    & Pressure & Climatology & Today \\ 
    & Pressure & Clim. Anomaly & Today \\ 
    & Pressure & Clim. Anomaly & Future \\ 
    \cline{2-4}
    & Wind (u) & Data & Today \\ 
    & Wind (u) & Data & Future \\ 
    \cline{2-4}
    & Wind (v) & Data & Today \\ 
    & Wind (v) & Data & Future \\  
    \cline{2-4}
    & Wind (module) & Data & Today \\ 
    & Wind (module) & Data & Future \\ 
    \hline
    \multirow{3}*{General}
    & Date (cos) & Data & Today \\ 
    & Date (sin) & Data & Today \\  
    \cline{2-4}
    & Land & Data & Today \\ 
    \hline
\end{tabular}

    \caption{Chosen configuration of the inputs for the experiments. ``Data'' in preprocessing means that no preprocessing except standardizing was performed. ``Past'' time interval means stacking all the specified maps for all the days in the past, including the last observable day (``Today''). ``Future'' --- stacking all the forecasts for the output days (3 for S-regime and 1 for R-regime). In R-regime corresponding forecasts from the last observable day replace data and forecasts of the coming days, so no yet unobserved data or forecasts leak to the model from the future.}
    \label{unet_configuration}
\end{table}

\subsection{Predicting Differences with a Baseline}
\label{section:difference_with_baseline}

The SIC map by itself is quite a complex image. Although U-Net architecture is well-suited for predicting local changes on an image, it may still struggle to reproduce the whole input, which may be similar to the desired output with some local changes, or require additional training time. In order to alleviate the problem for the model and to accelerate its training, we make it predict not the SIC data but the differences with a baseline. Since the persistence baseline performed best, we used it as the base $B$ and computed the model's forecast $F$ by the formula:
\begin{equation}
\label{base_equation}
    F = B + \alpha M.
\end{equation}
Here $M$ is the backbone's (U-Net) output, and we chose $\alpha = 0.1$ so that backbone outputs $M$ have approximately unit variance (which is the best for the default weights initialization \cite{kumar2017init}). We investigate the effect of this decision when we do ablations in subsection \ref{section:ablation_study}.

\subsection{Pretraining in R-regime}

In R-regime, we recurrently make next-day SIC forecast, using all the previous predictions as inputs (see subsection \ref{section:regimes}). The gradients flow through the backbone for $D_\text{out}$ times in this setting. We discovered that training from scratch in this regime proceeds very slowly and falls to non-optimal solutions that perform much worse than models trained in S-regime. The reason may be that R-regime is much more sensitive to proper initialization. To solve this problem, we divide the whole training into two parts:
\begin{enumerate}
    \item Pretraining the model in S-regime with $D_\text{out} = 1$;
    \item Initializing the model with the pretrained checkpoint and tuning it in R-regime for $D_\text{out}$ days.
\end{enumerate}
The pretraining is done as usual for 100 epochs, and tuning is done for just 20 epochs, which we found to be sufficient.

\subsection{3 Days Ahead Forecast}
\label{section:forecasting_for_3_days}

In this subsection we will describe results of the experiments conducted for $D_\text{in} = 7$ input days and $D_\text{out} = 3$ output days for the general model (trained on all three regions) with GFS data. This number of input days was chosen from these theoretical considerations: the model should require appropriate computational and memory resources for training yet be able to catch all the necessary trends and dynamics in the data. The number of output days is thought to be sufficient to investigate the model's forecast properties and compare the model's abilities in different settings. On the performance of the model with different numbers of input days see subsection \ref{section:ablation_study}, for the longer forecasts see subsection \ref{section:10_days_forecast}. All the metric values for our best models and baselines are collected in table \ref{metrics_table}.

\begin{table}[!htbp]
    \centering
    \begin{tabular}{llllll}
\toprule
 &  & Linear Trend & Persistence & U-Net (S) & U-Net (R) \\
Region & Metric &  &  &  &  \\
\midrule
\multirow[c]{3}{*}{Barents} & IIEE & 2.96 & 2.46 & 1.48 ± 0.02 & \textbf{1.41} ± 0.009 \\
 & MAE & 3.25 & 2.67 & 1.78 ± 0.01 & \textbf{1.73} ± 0.004 \\
 & RMSE & 9.8 & 8.44 & 5.68 ± 0.05 & \textbf{5.51} ± 0.05 \\
\midrule
\multirow[c]{3}{*}{Labrador} & IIEE & 1.82 & 1.54 & 0.905 ± 0.004 & \textbf{0.871} ± 0.01 \\
 & MAE & 1.66 & 1.41 & 0.966 ± 0.003 & \textbf{0.939} ± 0.009 \\
 & RMSE & 6.59 & 6.02 & 3.96 ± 0.03 & \textbf{3.88} ± 0.05 \\
\midrule
\multirow[c]{3}{*}{Laptev} & IIEE & 2.03 & 1.7 & 1.11 ± 0.03 & \textbf{1.05} ± 0.02 \\
 & MAE & 3.7 & 3.03 & 2.22 ± 0.02 & \textbf{2.16} ± 0.007 \\
 & RMSE & 8.87 & 7.61 & 5.1 ± 0.06 & \textbf{4.98} ± 0.05 \\
\bottomrule
\end{tabular}

    \caption{JAXA SIC metrics averaged over 3 forecast days and over 2021 for baselines and our best U-Net configurations (general with GFS). IIEE is computed for SIC classes with a 15\% binarization threshold. For the models, we report mean and unbiased std of 3 independent runs with random seeds 0, 1, and 2.}
    \label{metrics_table}
\end{table}

\begin{figure}[!htbp]
    \includegraphics[width=\textwidth]{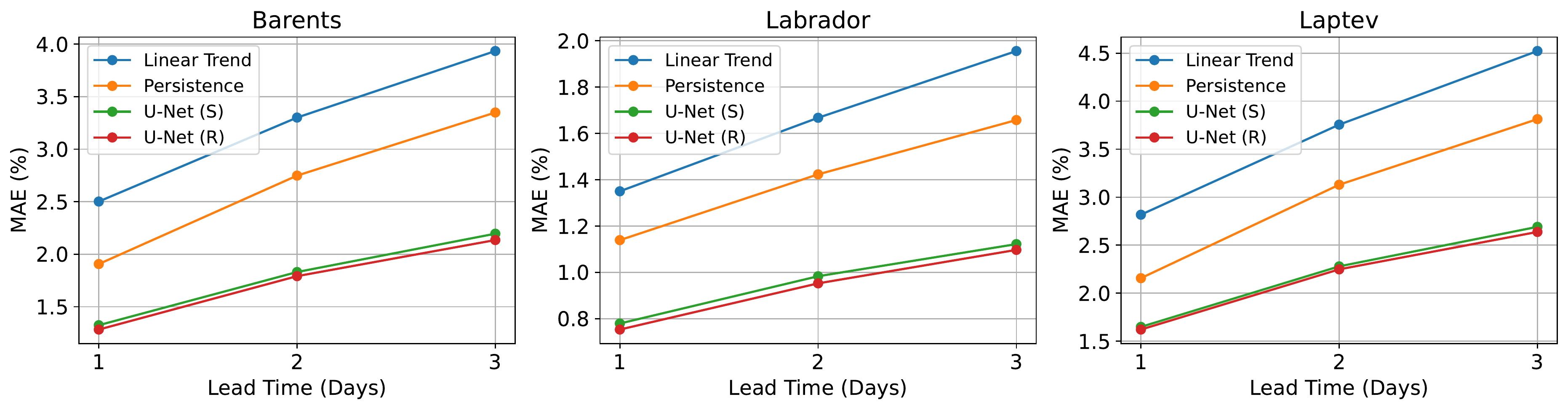}
    \centering
    \caption{Dependence of JAXA SIC MAE (lower is better) on the different forecast horizons (in days) for all three regions. MAE is averaged over the whole 2021 year. The linear trend is computed cell-wise over 3 previous days; U-Net (S) and U-Net (R) are trained on all three regions merged and shuffled, with 7 days history (past) and best inputs configuration, as presented in table \ref{unet_configuration}.}
    \label{jaxa_sic-models-mae-day}
\end{figure}

\begin{figure}[!htbp]
    \includegraphics[width=\textwidth]{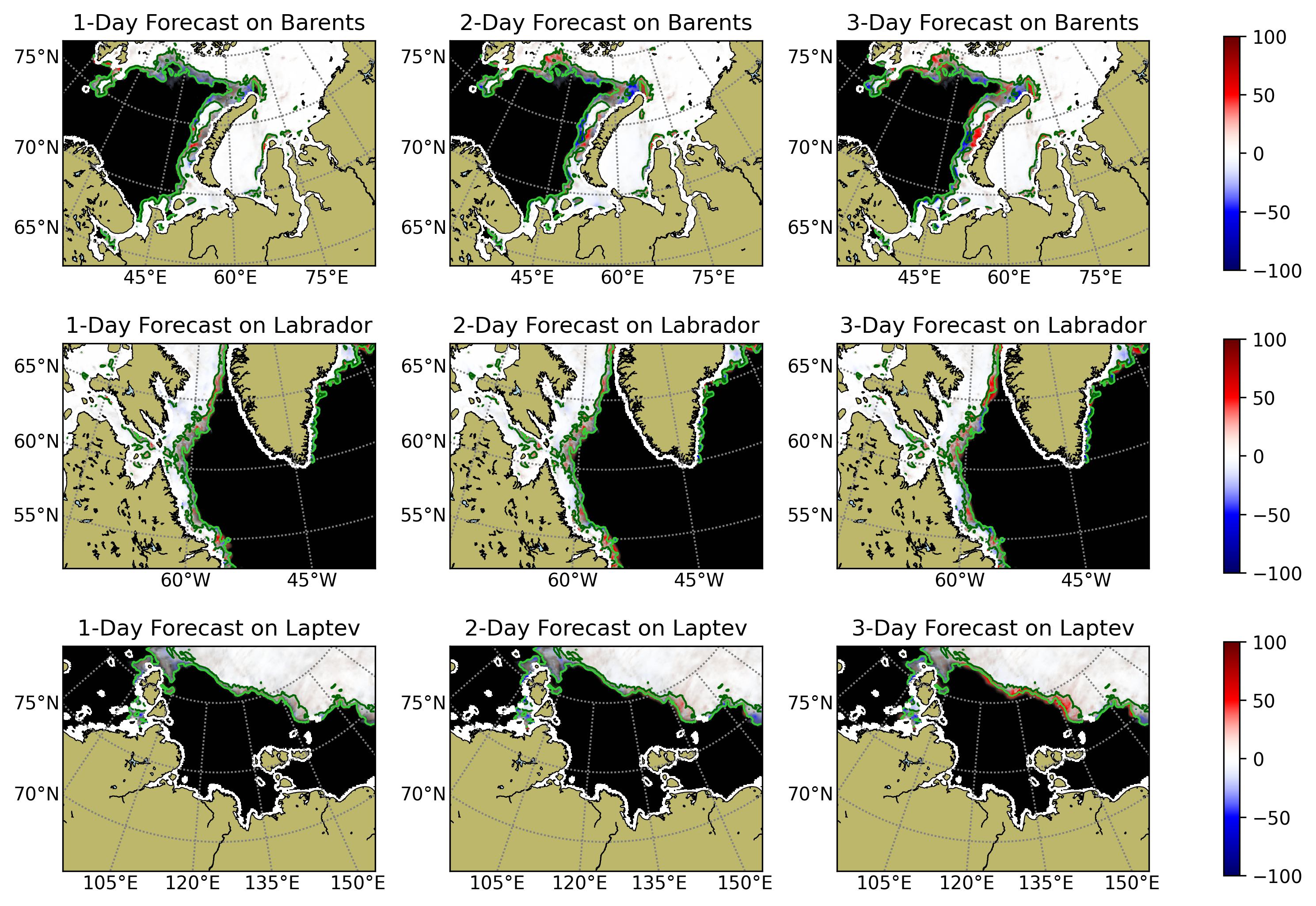}
    \centering
    \caption{U-Net (R) best configuration forecasts for fixed dates and varying forecast lead time. For Barents and Labrador, the date is 2021-04-01, and for Laptev, 2021-10-01, since on 2021-04-01, the region is almost frozen. Black-white color map shows the values of JAXA SIC, and the semitransparent red-blue color map shows the difference between model predictions and actual values of SIC. Light green isolines denote 15\% SIC, dark green --- 80\% SIC edge contouring the MIZ that lays in between. The relative size of MIZ for all three regions is the same and approximately 10\% of the sea area for the selected dates.}
    \label{jaxa_sic-forecasts-day-region}
\end{figure}

Since the forecasts with longer lead times (that are further in the future) are more challenging due to accumulating uncertainty, the error rate should increase as the number of lead-time days increases. This dependence is depicted in the figure \ref{jaxa_sic-models-mae-day}. U-Net in both regimes outperforms both baselines by a significant margin. Interestingly, linear trend performs noticeably worse than persistence, which we associate with high nonlinearity of the SIC dynamics in each cell along with measurement errors. U-Net (R) performs slightly better than U-Net (S) in all the cases, and the difference tends to increase with the forecast lead time. That might indicate that U-Net (R) model is more suitable for longer forecasts as it is trained in a fashion so to mitigate its errors when receiving its outputs as the next day's inputs. Examples of forecasts for fixed dates with 1, 2, or 3 lead-time days with the best configuration of U-Net (R) are presented in figure \ref{jaxa_sic-forecasts-day-region}. The green lines depicted in figure \ref{jaxa_sic-forecasts-day-region} contours the MIZ, and being overlaid with error (red-blue) shows that, like in numerical models, the most considerable discrepancies are contained inside it. One can see that most of the errors are accumulated near the edge of the ice, where most of the sea ice daily change takes place.

\begin{figure}[!htbp]
    \includegraphics[width=\textwidth]{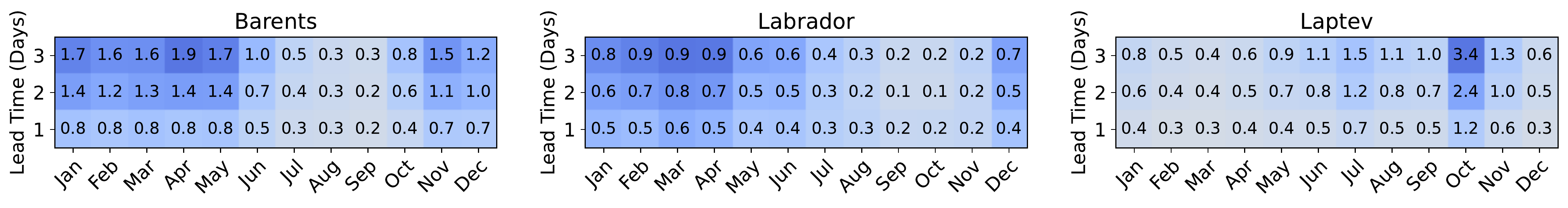}
    \centering
    \caption{Improvement of JAXA SIC MAE (in percentage points, higher is better) for general U-Net (S) with GFS over persistence for different months of 2021 and days of the forecast. All three regions are color-coded independently. Improvement is computed in absolute percentage points and tends to be higher for months with active sea ice change.}
    \label{jaxa_sic-persistence-unet_s-improvement-month}
\end{figure}

\begin{figure}[!htbp]
    \includegraphics[width=\textwidth]{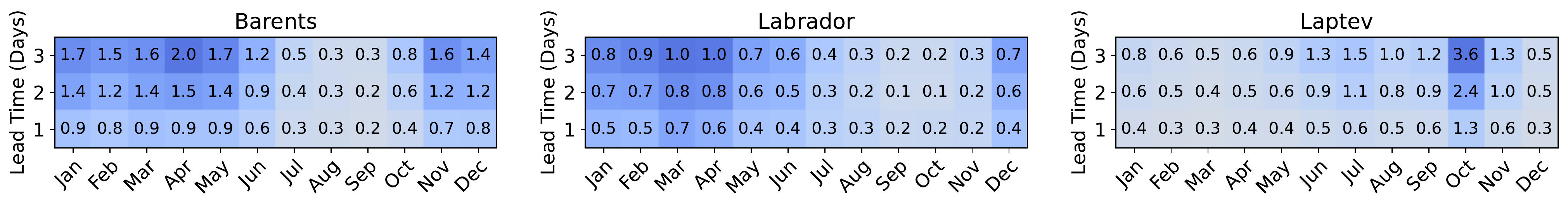}
    \centering
    \caption{Improvement of JAXA SIC MAE (in percentage points, higher is better) for general U-Net (R) with GFS over persistence for different months of 2021 and days of the forecast. All three regions are color-coded independently. Improvement is computed in absolute percentage points and tends to be higher for months with active sea ice change.}
    \label{jaxa_sic-persistence-unet_r-improvement-month}
\end{figure}

More informative are figures \ref{jaxa_sic-persistence-unet_s-improvement-month} and \ref{jaxa_sic-persistence-unet_r-improvement-month}, where absolute improvements of the model over persistence are presented separately for each month and color-coded. The similarity between Barents and Labrador regions, where the most improvement is obtained during winter and spring, and their distinction to the Laptev region, where the most improvement is obtained during summer and fall, are apparent.

\begin{figure}[!htbp]
    \includegraphics[width=\textwidth]{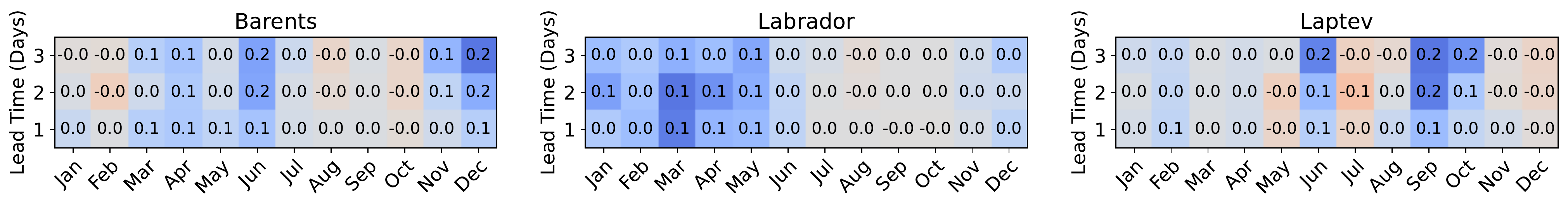}
    \centering
    \caption{Improvement of JAXA SIC MAE (in percentage points, higher is better) for general U-Net (R) with GFS over general U-Net (S) with GFS for different months of 2021 and different days of the forecast. All three regions are color-coded independently. Improvement is computed in absolute percentage points and tends to be higher for months with active sea ice change.}
    \label{jaxa_sic-unet_r-unet_s-improvement-month}
\end{figure}

\begin{figure}[!htbp]
    \includegraphics[width=\textwidth]{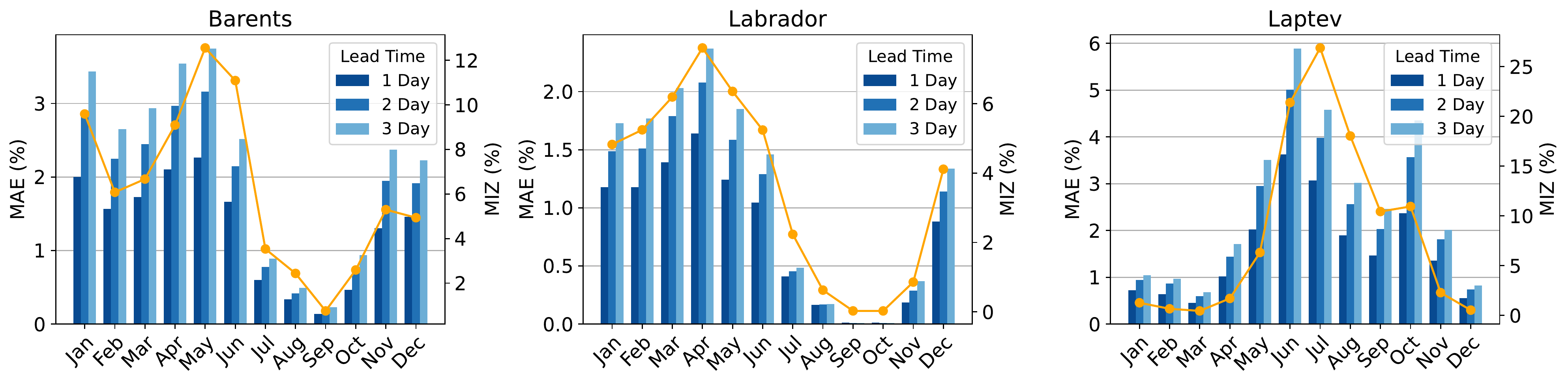}
    \centering
    \caption{General U-Net (R) with GFS: distribution of JAXA SIC MAE (lower is better) for different months of 2021 and days of the forecast. The orange line shows the marginal ice zone area relative to the region's sea area.}
    \label{jaxa_sic-mae-month}
\end{figure}

As was mentioned earlier, U-Net (R) performs slightly better than U-Net (S). This improvement is dissected in figure \ref{jaxa_sic-unet_r-unet_s-improvement-month}: U-Net (R) enhances the solution mostly during the generally challenging seasons: autumn/spring (due to icing/melting respectively) like it was with persistence baseline. Changing the ice regime from winter to summer is accompanied by enlarging the marginal ice zone - the area with the highest sea ice activity. This activity can be evaluated using statistical analysis of the distribution of cells' SIC climatological anomalies for each month, like those depicted in figures \ref{jaxa_sic-data-month} and \ref{jaxa_sic-anomaly-month}. Another way to estimate the impact of MIZ area on the solution quality is to compute the relative area (\%) of the marginal ice zone in the ocean domain (orange plot on figure \ref{jaxa_sic-mae-month}). In figure \ref{jaxa_sic-mae-month}, we see that the correlation between MAE and MIZ relative area almost equals 1. The high correlation means that even not rheology-dependent ML models face the same problem as numerical models \cite{Uotila2019}.
% Also, we note that the most considerable changes in the sea ice regime occur in Barents from November to May, in Labrador from December to May, and in Laptev from June to October. That follows the distribution of monthly errors of our best model represented in figure \ref{jaxa_sic-mae-month}.

% \begin{figure}[!htbp]
%     \includegraphics[width=\textwidth]{figures/jaxa_sic-miz-month-region.pdf}
%     \centering
%     \caption{Dependence of the marginal ice zone (MIZ) area to the sea area fraction on the calendar month in 2021 for all three regions. The higher values of MIZ indicate melting and freezing periods of sea ice with more complex sea ice dynamics.}
%     \label{jaxa_sic-miz-month-region}
% \end{figure}

\begin{figure}[!htbp]
    \includegraphics[width=\textwidth]{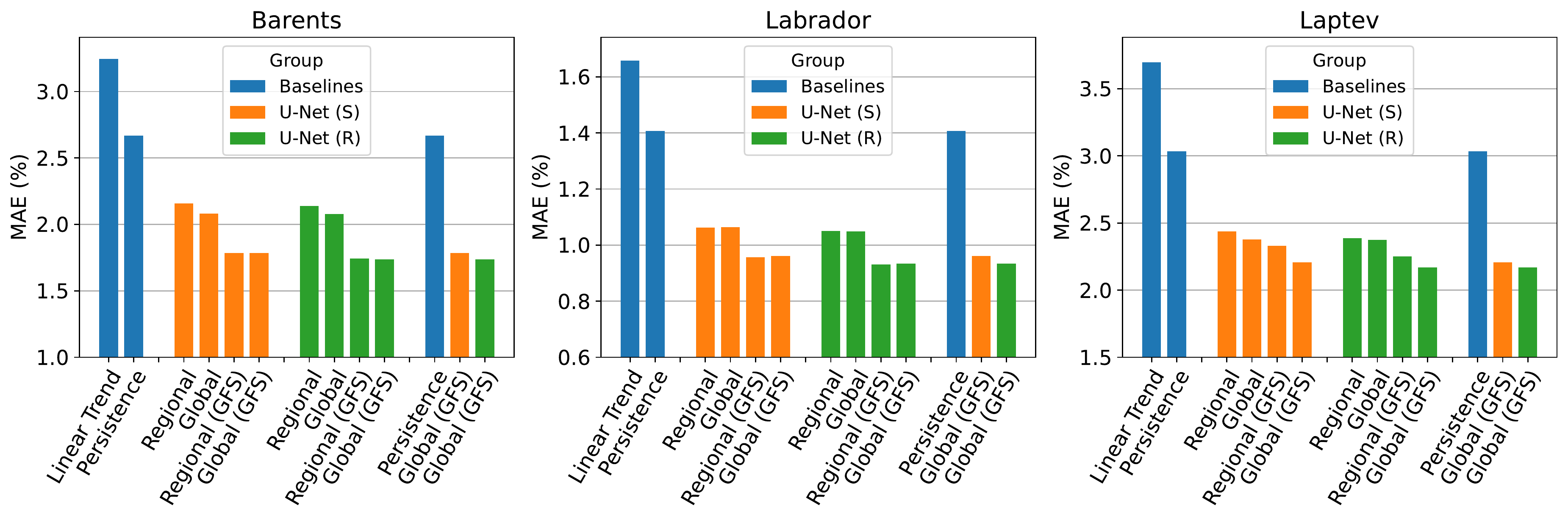}
    \centering
    \caption{JAXA SIC MAE averaged over 3 days of forecast and over 2021 (lower is better) for baselines and different configurations of U-Net. Colors denote model groups (baselines, U-Net (S), and U-Net (R)), configuration of each model within a group is specified below each bar. For U-Net regional and general configurations were trained with or without GFS. The best configurations from each group are presented separately on the right of each region plot for comparison.}
    \label{jaxa_sic-mae-model}
\end{figure}

\begin{figure}[!htbp]
    \includegraphics[width=\textwidth]{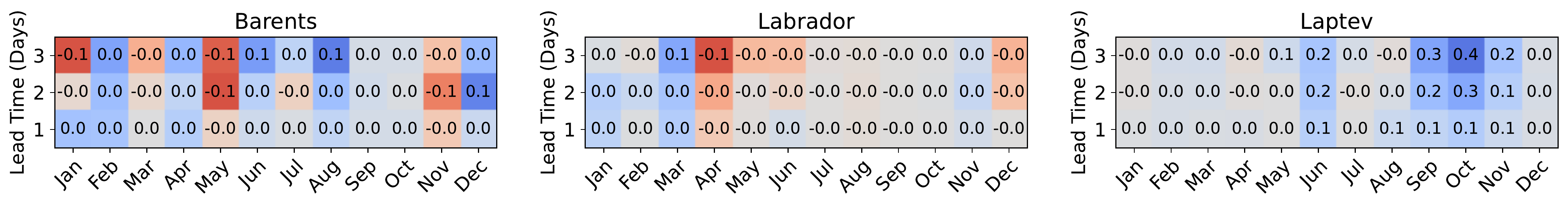}
    \centering
    \caption{Improvement of JAXA SIC MAE (in percentage points, higher is better) for general U-Net (R) with GFS over regional U-Net (R) with GFS for different months of 2021 and different days of the forecast. All three regions are color-coded independently. Improvement is computed in absolute percentage points and tends to be higher for months with active sea ice change.}
    \label{jaxa_sic-regional-improvement-month}
\end{figure}

\begin{figure}[!htbp]
    \includegraphics[width=0.5\textwidth]{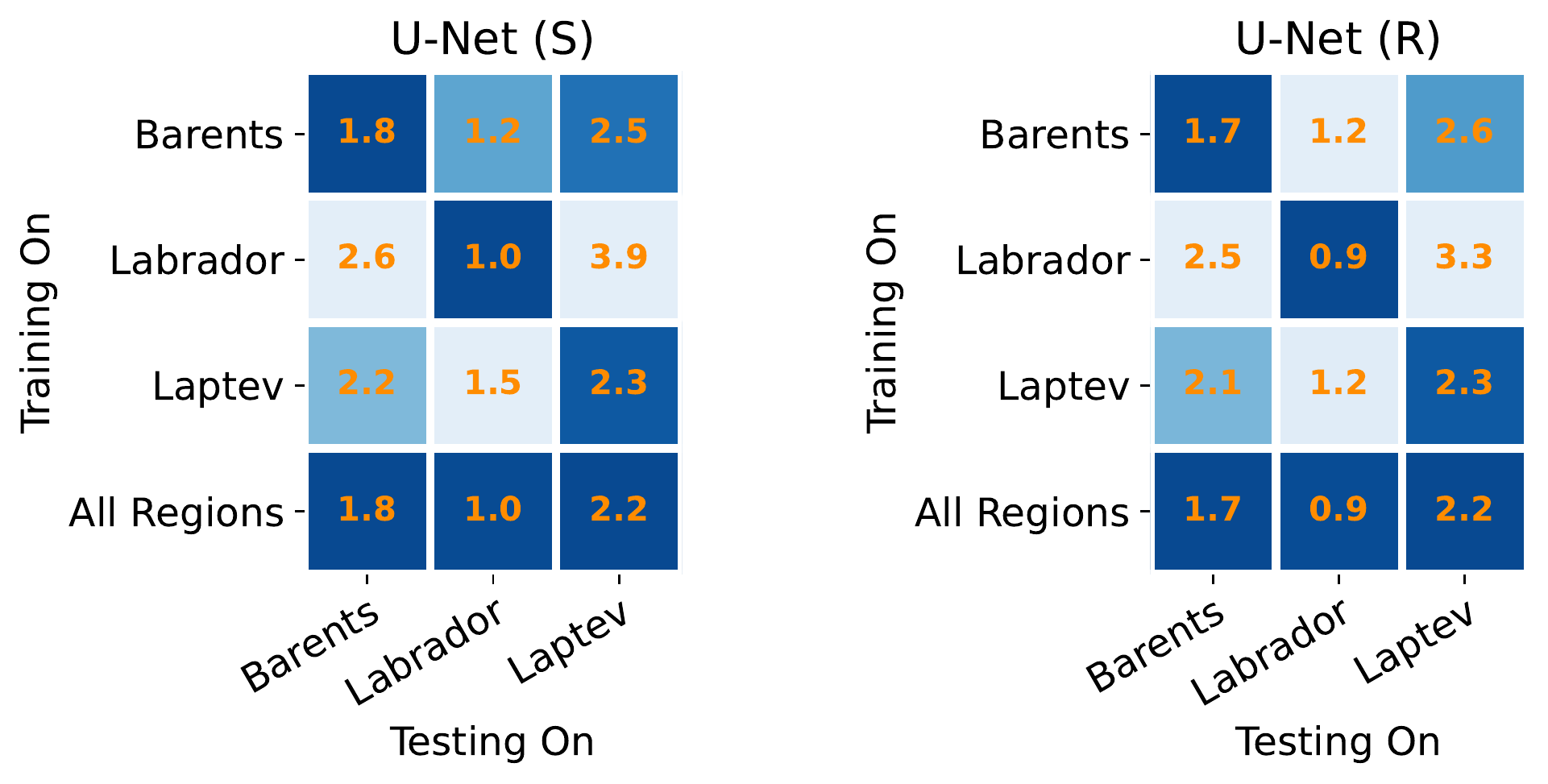}
    \centering
    \caption{JAXA SIC MAE averaged over 3 days of forecast and over 2021 (lower is better) for different train-test region configurations. The models trained in any region are expected to perform best in this region, and the general model (trained in all three regions) is expected to perform well in all of them. Each column (testing region) color-coded separately results in a dark blue diagonal matrix with a dark blue last row.}
    \label{jaxa_sic-region-region}
\end{figure}

All the results above are given for the general models. That means that both U-Net (S) and U-Net (R) were trained on three regional datasets merged in one and shuffled. One can expect that more diverse data will be helpful for the model to perform better. To demonstrate it, we conducted experiments and compared the general model with all three regional ones in both S- and R-regimes. The performances of all the settings are represented in figure \ref{jaxa_sic-mae-model}. The most boost is achieved by including GFS fields in the inputs; switching between general and regional settings helps most in the Laptev region, does not change in the Barents region, and gives a nonsignificant decrease in performance in the Labrador region. The change when switching from regional models to the general one is dissected in figure \ref{jaxa_sic-regional-improvement-month} for U-Net (R) (for U-Net (S), it looks almost the same). In figure \ref{jaxa_sic-region-region} regional models are also tested on the other two unseen regions and compared to the general model. In both cases, the upper $3 \times 3$ square diagonal is dark blue because the models are tested on their native regions, as is the bottom row with general model results. We prefer the general model to regional ones because of its universality and expect it to generalize better to previously unseen regions.

\subsection{10 Days Ahead Forecast}
\label{section:10_days_forecast}

\begin{figure}[!htbp]
    \includegraphics[width=\textwidth]{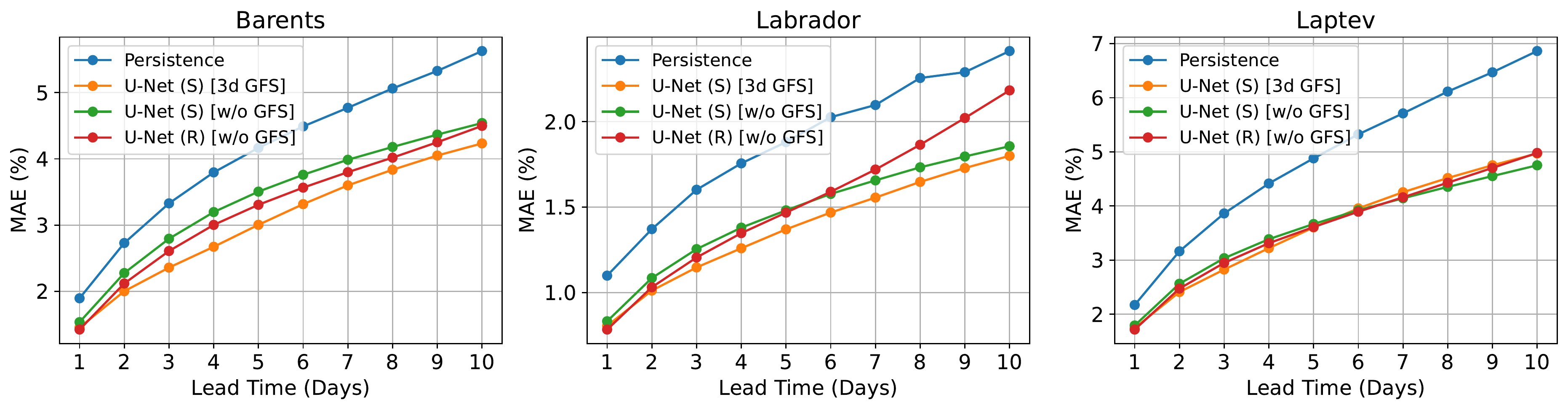}
    \centering
    \caption{Dependence of JAXA SIC MAE (lower is better) on the number of forecasted day in the future (lead time of the forecast) for all three regions. MAE is averaged over the whole 2021 year. U-Net (S) and U-Net (R) are trained on all three regions merged and shuffled, with 7 days history (past) and best inputs configuration, as presented in table \ref{unet_configuration}.}
    \label{jaxa_sic-long-mae-day}
\end{figure}

\begin{figure}[!htbp]
    \includegraphics[width=\textwidth]{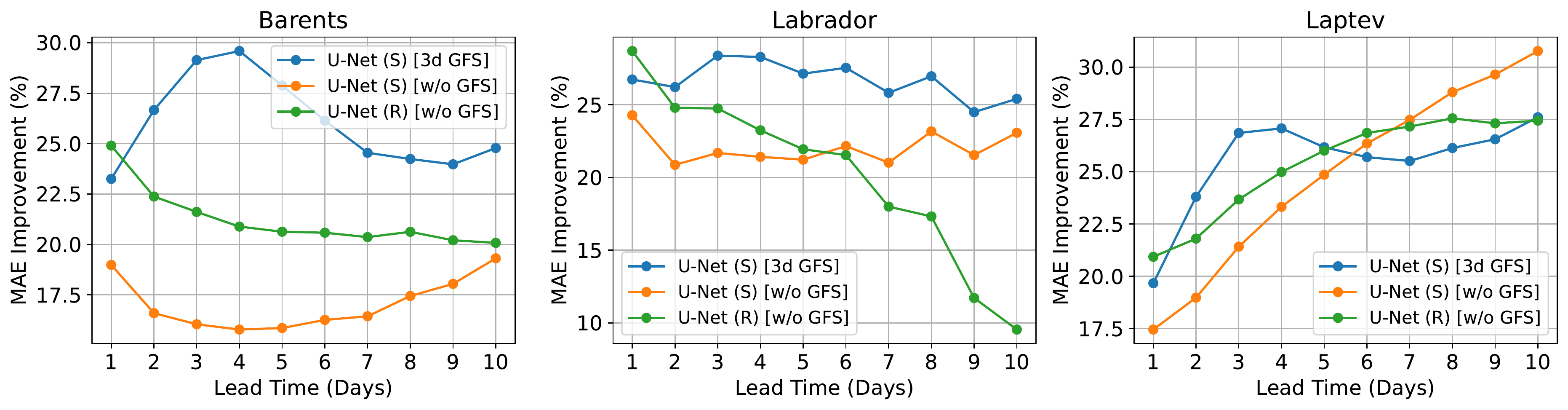}
    \centering
    \caption{Dependence of JAXA SIC MAE relative improvement over persistence (higher is better) on the different forecast horizons (in days) for all three regions. MAE is averaged over the whole 2021 year. U-Net (S) and U-Net (R) are trained on all three regions merged and shuffled, with 7 days history (past) and best inputs configuration, as presented in table \ref{unet_configuration}.}
    \label{jaxa_sic-long-improvement-day}
\end{figure}

We also trained U-Net (S) to make forecast for $D_\text{out} = 10$ days and tested U-Net (R) trained with $D_\text{out} = 3$ for 10 output days. We did not train U-Net (R) with $D_\text{out} = 10$ days because the memory and computational requirements, in this case, were too big. We also trained U-Net (R) model without GFS channels since we only had GFS forecasts for the next 3 days. The results are depicted in figure \ref{jaxa_sic-long-mae-day}. To better understand the influence of the presence of GFS channels in the inputs, we also conducted experiments for U-Net (S) with GFS data and computed relative improvement over persistence for all three settings. The results are presented in figure \ref{jaxa_sic-long-improvement-day}. One can see the improvement of 5\%-15\% for the first 3-4 days for the U-Net (S) with the GFS setting in Barents and Laptev compared to U-Net (S) without the GFS setting. For Labrador, the improvement is smaller but persists for all 10 days of forecast, as it does for Barents. U-Net (R) without GFS generally performs better for the first half of the forecast days than U-Net (S) without GFS. Its performance deteriorates as the forecast lead time increases. This fact is under general knowledge about RNNs, which performance usually deteriorates when the depth of recurrence increases.

\subsection{Ablation Studies}
\label{section:ablation_study}

\begin{figure}[!htbp]
    \includegraphics[width=\textwidth]{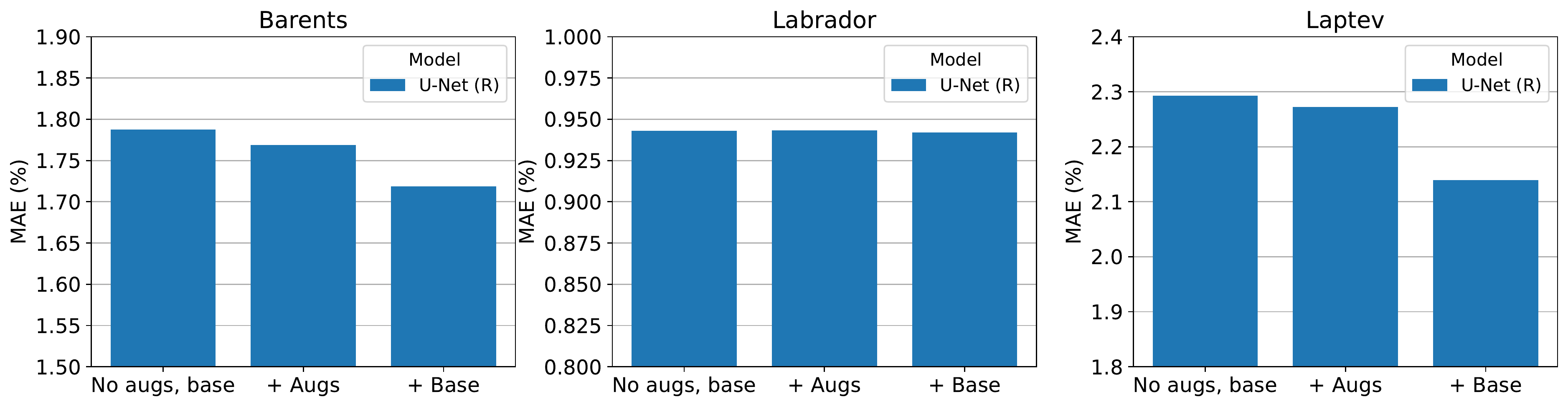}
    \centering
    \caption{JAXA SIC MAE averaged over 3 forecast days and over 2021 for different ablations of general U-Net (R) configuration with GFS. From left to right on each diagram, we first turn on augmentations and then add persistence base to the predictions (as in equation \ref{base_equation}) and report the model's metric.}
    \label{jaxa_sic-mae-ablation}
\end{figure}

In subsection \ref{section:augmentations} we introduced the set of augmentations we used. In subsection \ref{section:difference_with_baseline} we described how we predict not the raw SIC data but the difference with the persistence baseline to alleviate the problem for U-Net. In figure \ref{jaxa_sic-mae-ablation} we demonstrate the influence of both these factors on the model's accuracy for our best configuration (general U-Net (R) with GFS). They boost MAE in Barents and Laptev regions and do not make any significant difference in the Labrador region.

\begin{figure}[!htbp]
    \includegraphics[width=\textwidth]{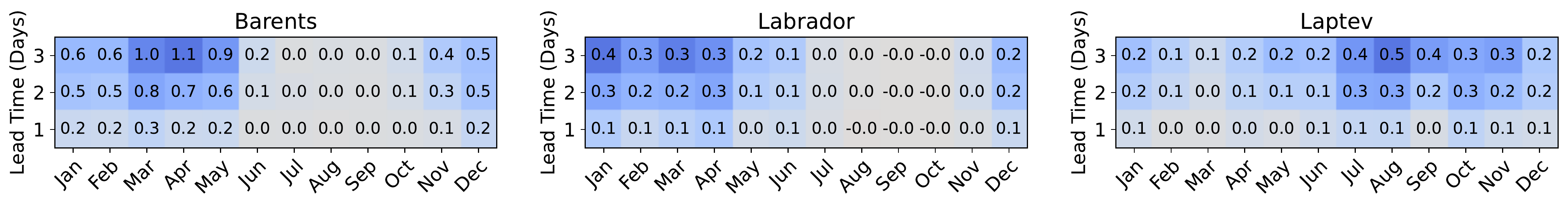}
    \centering
    \caption{Improvement of JAXA SIC MAE (in percentage points, higher is better) for general U-Net (S) with GFS over general U-Net (S) without GFS for different months of 2021 and different forecast horizons (in days). All three regions are color-coded independently. Improvement is computed in absolute percentage points and tends to be higher for months with active sea ice change.}
    \label{jaxa_sic-unet_s-gfs-improvement-month}
\end{figure}

\begin{figure}[!htbp]
    \includegraphics[width=\textwidth]{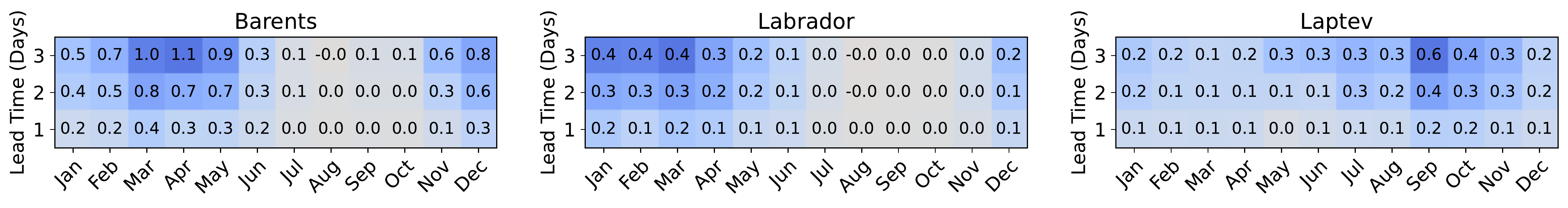}
    \centering
    \caption{Improvement of JAXA SIC MAE (in percentage points, higher is better) for general U-Net (R) with GFS over general U-Net (R) without GFS for different months of 2021 and different forecast horizons (in days). All three regions are color-coded independently. Improvement is computed in absolute percentage points and tends to be higher for months with active sea ice change.}
    \label{jaxa_sic-unet_r-gfs-improvement-month}
\end{figure}

We also investigated the influence of including the GFS channels in the model and dissected the improvement over months. The results are presented in figures \ref{jaxa_sic-unet_s-gfs-improvement-month} and \ref{jaxa_sic-unet_r-gfs-improvement-month}, and in both cases, these channels are proved to be very useful for the model, especially during the months with an active change of sea ice.

\section{Discussion}
\label{section:discussion}

In short-term predictions (for 3 days), U-Net (R) slightly outperformed U-Net (S), and both of them outperformed baselines by a significant margin (table \ref{metrics_table}), demonstrating high prospects of machine learning methods in sea ice forecasting. For longer lead times (10 days ahead forecasts), U-Net (R) quality deteriorates as anticipated, and it usually yields to U-Net (S) starting from the second half of the forecast (figure \ref{jaxa_sic-long-improvement-day}). Extra weather forecast channels improve forecasts quality noticeably for the days when the weather forecast is provided and to some extent for subsequent days (figure \ref{jaxa_sic-long-improvement-day}). The marginal ice zone stands as the most challenging part of a region, and the months of most active ice change stand as the most challenging part of a year for forecasting (figures \ref{jaxa_sic-forecasts-day-region} and \ref{jaxa_sic-mae-month}). Finally, training not only on the region of interest but also on more diverse data improves the overall performance of the model and its generalization abilities (figure \ref{jaxa_sic-region-region}).

We utilized U-Net architecture for our experiments, and it performed well. It is lightweight, thus not prone to overfitting, yet suited well for image-to-image tasks, such as sea ice forecasting. However, there are a few more specialized neural network architectures used mainly in video prediction task (PredRNNs \cite{wang2017predrnn, wang2018predrnnpp, wang2021prernnv2}, E3D-LSTM \cite{wang2019eid3dlstm}, CrevNet \cite{yu2020crevnet}), that is closely related to sea ice forecasting. One possible direction for future work can be embedding these backbones into the pipeline and comparing their performance. Another important extension is to compare the performance of numerical ocean-ice models, such as GLORYS12-V1, TOPAZ4, or SODA3.3.1, with the developed data-driven approach. \cite{Uotila2019} showed that the ten most modern ocean reanalyses systematically underestimate the area of MIZ during spring and autumn even with data assimilation. Finally, one can try to fuse these approaches and train statistical models to compensate for errors of the numerical models or use them in any other more intricate way. These may include more flexible sparse data assimilation (e.g., from buoys) or incorporating physics into data-driven models via NeuralODE approaches \cite{chen2018neuralode}. Adding extra data, particularly one that represents types of sea ice and its thickness, may also significantly boost the forecasting quality. The main difficulty here may be to locate reliable and complete sources of these operative data.

\section{Conclusions}

Data-driven models based on machine learning are gaining popularity as fast and robust alternatives for numerical ocean-ice models in short-range weather forecasting. We investigate their efficiency in sea ice forecasting in several Arctic regions. First, we collect JAXA AMSR-2 Level-3 SIC data and GFS analysis and forecasts data, process it and construct three regional datasets, which can be used as benchmark tasks in future research. Second, we conduct numerous experiments on forecasting SIC maps with the U-Net model in two regimes and provide our findings on the prospect of this approach, including comparison with standard baselines, standard metric values, and model generalization ability. That allows us to build a fast and reliable tool --- trained on all three regions U-Net network, that can provide operational sea ice forecasts in any Arctic region. Finally, we compare U-Net forecasting performance in recurrent (R) and straightforward (S) regimes and highlight the strengths and weaknesses of both these regimes.

\section*{Supplementary Materials}
The following supporting information can be downloaded at:  \url{https://disk.yandex.ru/d/n3PaW0p04GPiwg}, Video S1: barents-1d-forecasts.mp4, Video S2: barents-2d-forecasts.mp4, Video S3: barents-3d-forecasts.mp4, Video S4: labrador-1d-forecasts.mp4, Video S5: labrador-2d-forecasts.mp4, Video S6: labrador-3d-forecasts.mp4, Video S7: laptev-1d-forecasts.mp4, Video S8: laptev-2d-forecasts.mp4, Video S9: laptev-3d-forecasts.mp4.

\section*{Funding}
The work was supported by the Analytical center under the RF Government (subsidy agreement 000000D730321P5Q0002, Grant No. 70-2021-00145 02.11.2021).

\section*{Data Availability}
Interpolated and preprocessed data for all three regions (Barents, Labrador, Laptev) used in this study are openly available at \url{https://disk.yandex.ru/d/5Qc_OhbU7NYQew}. See README.md there for the details.

\section*{Acknowledgments}
We gratefully appreciate insightful discussions with experts from Gazprom Neft and Avtomatika Service and thank the Association ``Artificial intelligence in industry'' for providing a platform for such discussions.

\appendix
% \section[\appendixname~\thesection]{}

% \subsection[\appendixname~\thesubsection]{}

% The appendix is an optional section that can contain details and data supplemental to the main text---for example, explanations of experimental details that would disrupt the flow of the main text but nonetheless remain crucial to understanding and reproducing the research shown; figures of replicates for experiments of which representative data are shown in the main text can be added here if brief, or as Supplementary Data. Mathematical proofs of results not central to the paper can be added as an appendix.

% \section[\appendixname~\thesection]{}

% All appendix sections must be cited in the main text. In the appendices, Figures, Tables, etc. should be labeled, starting with ``A''---e.g., Figure A1, Figure A2, etc.

%Bibliography
\bibliographystyle{unsrt}
\bibliography{references}

\end{document}